\documentclass[journal]{IEEEtran}

\usepackage{comment} 
\usepackage{amsmath,amssymb,bm}
\usepackage{subcaption,graphicx}
\usepackage[percent]{overpic}
\usepackage{cite}
\usepackage{changes}
\usepackage[textsize=footnotesize]{todonotes} 
\usepackage{placeins}

\definechangesauthor[name={Markku}, color=orange]{m}
\definechangesauthor[name={Smutny}, color=red]{vs}

\makeatletter
\def\endthebibliography{%
\def\@noitemerr{\@latex@warning{Empty `thebibliography' environment}}%
\endlist
}
\makeatother

\title{Static Stability of Robotic Fabric Strip Folding}

\author{Vladim\'{i}r~Petr\'{i}k, \and Vladim\'{i}r~Smutn\'y,~\IEEEmembership{Member,~IEEE,} \and Ville~Kyrki,~\IEEEmembership{Senior~Member,~IEEE}%
\thanks{%
This work was supported by Academy of Finland, decision 317020.
}
\thanks{%
Vladim\'{i}r Petr\'{i}k and Ville Kyrki are with Department of Electrical Engineering and Automation, Aalto University, Finland. E-mails: \{vladimir.petrik, ville.kyrki\}@aalto.fi}
\thanks{
Vladim\'{i}r Smutn\'y is with Czech Institute of Informatics, Robotics, and Cybernetics, Czech Technical University in Prague. E-mail: vladimir.smutny@cvut.cz}%
}

\newcommand{\etamunit}{$\mathrm{m^{-2} \, s^{2}}$}

\begin{document}

    \maketitle
    \IEEEpeerreviewmaketitle

    \begin{abstract}
    Planning accurate manipulation for deformable objects requires prediction of their state.
    The prediction is often complicated by a loss of stability that may result in collapse of the deformable object.
    In this work, stability of a fabric strip folding performed by a robot is studied.
    We show that there is a static instability in the folding process.
    This instability is detected in a physics-based simulation and the position of the instability is verified experimentally by real robotic manipulation.
    Three state-of-the-art methods for folding are assessed in the presence of static instability.
    It is shown that one of the existing folding paths is suitable for folding of materials with internal friction such as fabrics.
    Another folding path that utilizes dynamic motion exists for ideal elastic materials without internal friction.
    Our results show that instability needs to be considered in planning to obtain accurate manipulation of deformable objects.
\end{abstract}

    \section{INTRODUCTION}\label{sec:introduction}
\IEEEPARstart{P}{lanning} robotic manipulation of deformable objects remains a challenging research area.
The examples of deformable objects include flexible cables, ropes, elastic or paper sheets, or fabrics.
These objects have a large or infinite number of degrees of freedom~\cite{sanchez2018robotic}.
This fact complicates the prediction of the object state, for which a complex dynamic model is needed.
Compared to rigid bodies, deformable objects are more difficult to immobilize by conventional grippers.
Due to the large number of degrees of freedom, some of them often remain free, and the system may become autonomous which results in an uncontrollable motion or collapse.
Special treatment needs to be considered for planning in the state space where the system becomes autonomous.

\begin{figure}[t]
    \centering
    \subcaptionbox{Initial state}[.8\linewidth]{\includegraphics[width=.8\linewidth]{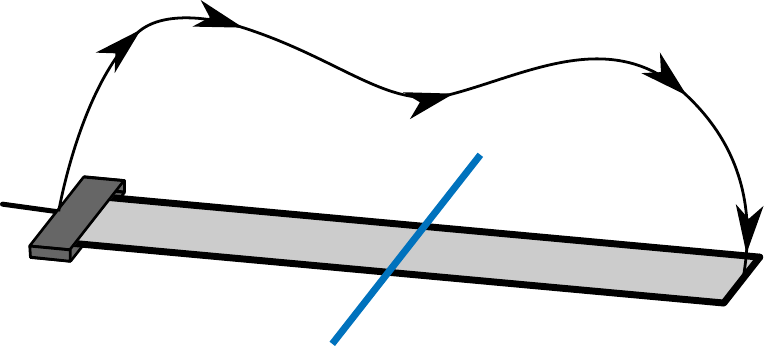}}\\
    \subcaptionbox{Folded state}[.8\linewidth]{\hfill\includegraphics[width=.5\linewidth]{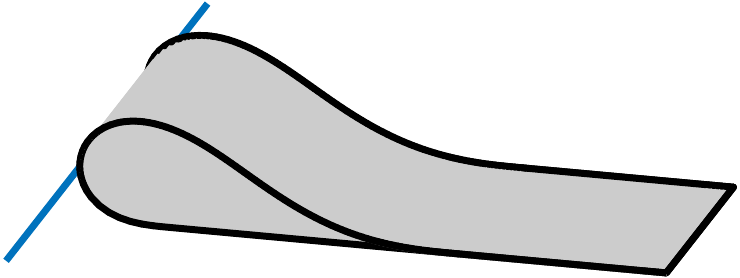}}%
    \caption{
    Folding of a fabric strip parametrized by a folding line (blue line).
    Figure~(a) shows an initial state which is grasped by a gripper at one side.
    The folding path (black arrows) is followed by the gripper.
    The optimal folding path resembles the expected folded state~(b).
    }
    \label{fig:intro}
\end{figure}

One example of deformable object manipulation task is garment folding.
In a folding process, a robot picks the garment at one side and, by following the folding path, brings the garment from its flat initial configuration to the expected folded shape~(Fig.~\ref{fig:intro}).
In recent years, researchers have focused on the planning of a folding path which provides an accurately folded garment~\cite{LiIROS2015,PetrikADR2017}.
Good folding accuracy has been achieved by using a simulation of the garment and by analysing its shape when performing the folding.
However, the autonomous behavior of the garment in the folding process and its effect on the folding accuracy have not been analyzed before.

This paper provides an analysis of such an autonomous behavior in the single layer fabric strip folding.
It is shown that there is a static instability in the folding process.
This instability results in a dynamic motion which is not controllable by a robot and corresponds to the autonomous behavior of the model.
The static instability is firstly analyzed for a simple motion in a physics-based simulation.
The states of the strip where instability occurs in the simulation are compared to real fabric strip measurements.
Then folding paths proposed in literature are analyzed in simulation and implications of static instability for the folding planning are discussed.

The contributions of the paper are:
\begin{itemize}
    \item We show that there is a static instability in the fabric strip folding.
    \item We demonstrate how this instability can be detected using a physics-based simulation and compare it to physical measurements.
    \item We analyze how the existing methods deal with the presence of the instability and discuss implications of our observations on the folding planning.
    \item We show that internal friction is required to fold an elastic strip statically.
\end{itemize}

    \section{STATE OF THE ART}\label{sec:stateOfTheArt}

\begin{figure*}[t]
    \centering
    \subcaptionbox{\textit{Triangular path}}[.33\linewidth]{\includegraphics[width=.32\linewidth]{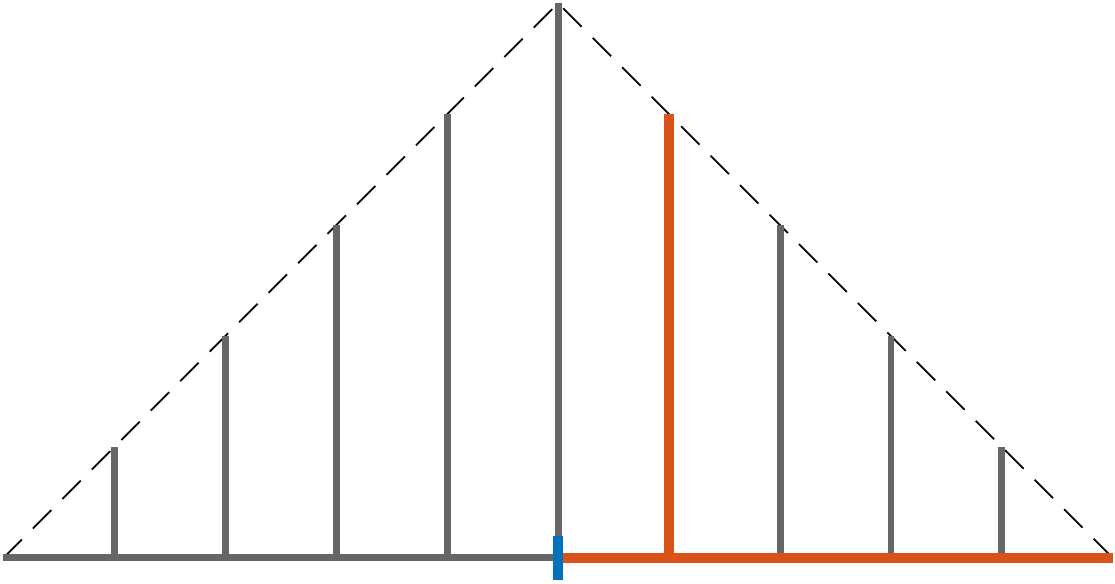}}%
    \subcaptionbox{\textit{Circular path}}[.33\linewidth]{\includegraphics[width=.32\linewidth]{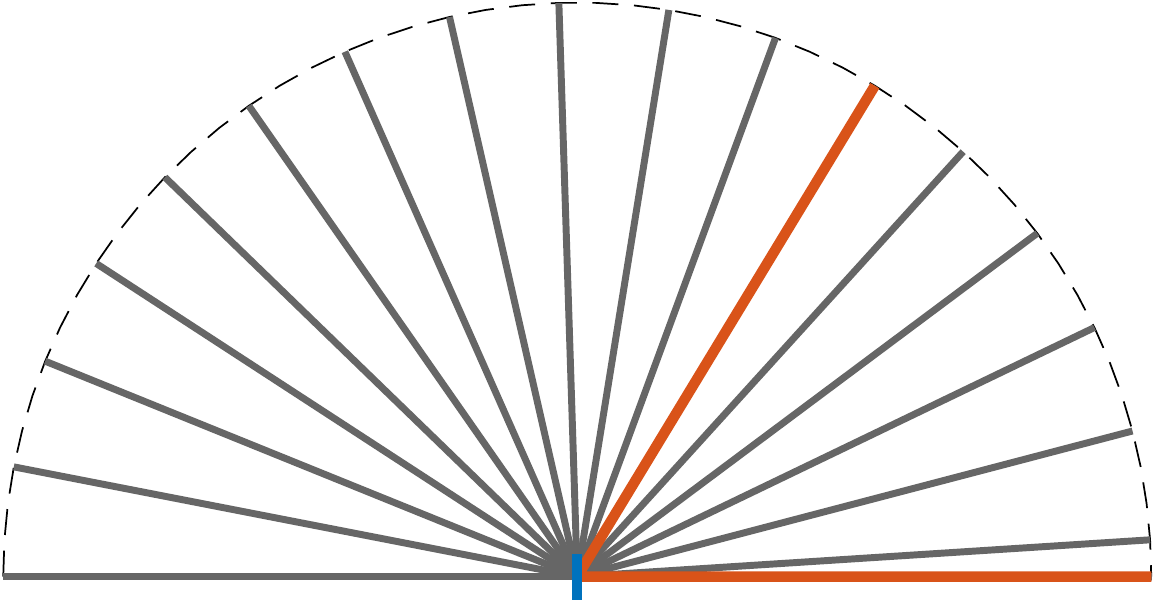}}%
    \subcaptionbox{\textit{R-path}}[.33\linewidth]{\includegraphics[width=.32\linewidth]{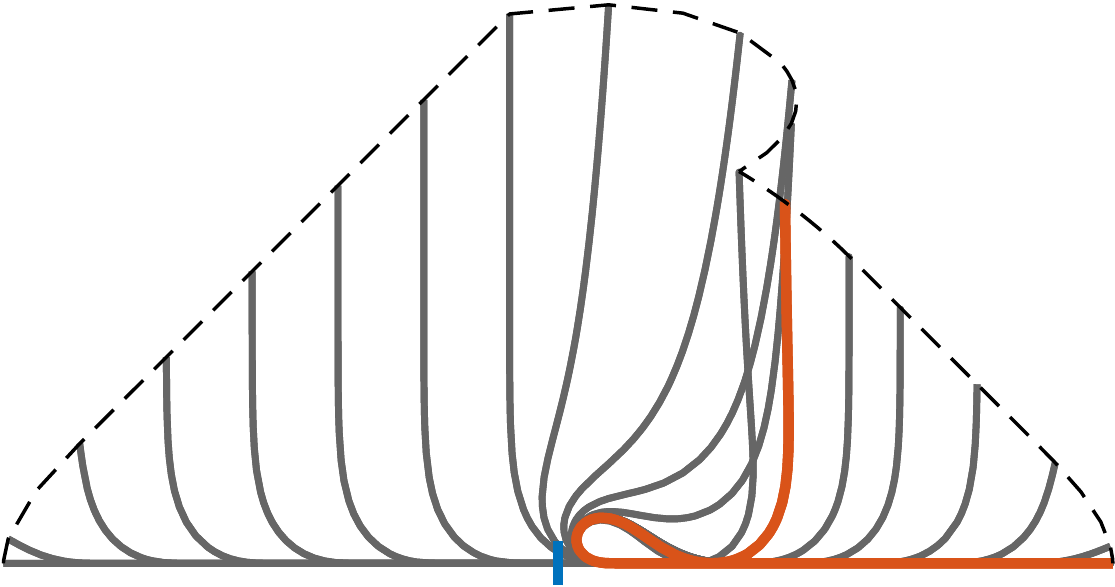}}%
    \caption{
    Different folding paths (dashed line) and the states (solid lines) assumed in the planning of the path.
    The blue marker shows position of the folding line.
    The \textit{triangular path}~(a) assumes infinitely flexible garment.
    The \textit{circular path}~(b) assumes rigid material with flexibility in the folding line only.
    The \textit{R-path}~(c) models flexibility by a single parameter.
    }
    \label{fig:sotapaths}
\end{figure*}

The study of the garment behaviour is necessary for predicting the garment state accurately.
The analysis of the behaviour was performed for various tasks including garment drying~\cite{drying}, flattening and ironing~\cite{LiSunTAROS2013,LiSunICRA2015,YinxiaoICRA2016,EstevezIROS2017}, or folding~\cite{MillerIJRR2011}.
A garment manipulation is analysed in our study.

A garment manipulation pipeline can be considered to consist of four tasks: (a)~isolating, (b)~unfolding and classifying, (c)~folding, and (d)~stacking of the folded piece of garment~\cite{HamajimaRAS2000}.
The goal of the isolating task is to pick a single piece of garment from a heap of garments.
The main parts of the task are the detection of a point to grasp and the grasping itself.
Several strategies for grasping were compared in~\cite{AlenyaICRA2012}.
Furthermore, specialized grippers have been proposed for handling garments in~\cite{HamajimaRAS2000,LeClopemaGripper,zhang1999modeling}.
Unfolding and classification task~\cite{CusumanoTownerICRA2011,DoumanoglouICRA2014,TriantafyllouRAS2016} ensures that the garment lies flattened on a folding surface prior to folding.
The stacking task cleans the folding surface after the folding is done.
All tasks were successfully performed for a towel in~\cite{Maitin-ShepardICRA2010}.
Performing the pipeline for a wider range of garment types was proposed in~\cite{DoumanoglouTRO2016,Yinxiao2017}.

In the folding task, the garment lies flattened on a surface and the garment's type is known.
The folding can be decomposed to several independent folds based on the garment type as suggested in~\cite{MillerIJRR2011}.
In~\cite{MillerIJRR2011}, the robot performed folding in an open-loop manner.
A more robust version was proposed in~\cite{StriaIROS2014} where the garment pose was detected after each fold.
Both works~\cite{MillerIJRR2011,StriaIROS2014} parametrized an individual fold by grasping points and a folding line.
The folding line divides the garment into two parts such that at the end of the folding one part lies on the other part~(Fig.~\ref{fig:intro}).

In each fold, a robot is following a folding path.
Several methods for designing the folding path have been proposed.
The methods could be divided into several categories according to the complexity of the path computation:
(i) the geometrical methods (Sec.~\ref{subsec:geometricalPaths}) design a path which depends on the garment dimensions only;
(ii) the simulation based methods perform an underlying static (Sec.~\ref{subsec:staticSimulationBasedPath}) or dynamic (Sec.~\ref{subsec:dynamicSimulationBasedPath}) simulation to design a path;
and (iii) the learning based methods (Sec.~\ref{subsec:learningBasedApproaches}) learn a feedback control policy from observations.

In this work, we analyse how the geometrical and static simulation based paths deal with the instability in the folding process.
Other state-of-the-art methods require specialised hardware to fold the garment in a closed-loop manner or their underlying model is not comparable to our finite element simulation.
However, we list these methods to make the overview of the folding path generation complete.

\subsection{Geometrical Paths}\label{subsec:geometricalPaths}
Two folding paths which depend on the garment dimensions only were proposed in literature: the \textit{triangular}~\cite{BergWAFR2010} and \textit{circular}~\cite{PetrikTAROS2015}.
The \textit{triangular path} was designed based on assumptions that the garment is infinitely flexible and the friction between the garment and the desk is infinite.
The \textit{circular path} assumed a rigid material with flexibility in folding line only.
Both paths are visualised in Fig.~\ref{fig:sotapaths}.
In this paper we analyse how both paths deal with the static instability in a fabric folding.

\subsection{Static Simulation Based Path}\label{subsec:staticSimulationBasedPath}
A physics-based simulation was used in~\cite{PetrikIROS2016} to simulate a fabric strip according to \textit{Euler-Bernoulli} beam theory.
A fabric material was parametrized by a single parameter~\textit{weight-to-stiffness-ratio}.
The folding path was derived based on the static equilibrium of forces.
The method was extended to single arm rectangular fabric sheets folding in~\cite{PetrikADR2017}.
A \textit{Kirchoff-Love} shell theory was used for model and the material was parametrized by three parameters \textit{weight-to-membrane-stiffness-ratio}, \textit{weight-to-bending-stiffness-ratio}, and Poisson’s ratio.
In~\cite{PetrikICRA2018} it was shown that \textit{weight-to-bending-stiffness-ratio} is sufficient to describe the fabric and the method for estimating this parameter was proposed.
Path designed according to this method is visualised in Fig.~\ref{fig:sotapaths}.
As its shape resembles `R' letter it is called \textit{R-path}, hereinafter.
In this work, we model a fabric strip as a 3D elastic solid instead of using beam~\cite{PetrikIROS2016} or shell theory~\cite{PetrikADR2017}.
The parameters of the state-of-the-art method~\cite{PetrikADR2017} are related to our model.
It allows us to analyze this folding path in the presence of static instability.

\subsection{Dynamic Simulation Based Path}\label{subsec:dynamicSimulationBasedPath}
A specialised high-speed robot hands were used in a series of works~\cite{yamakawa2010dynamic,yamakawa2011dynamic,yamakawa2011motion} to perform a dynamic-based folding of a towel.
The hands swipe the towel in air according to the motion planned on a simulated model.
The finger of the hands catch the swiping end of the towel based on the camera feedback.
A specialised hardware is required to perform this dynamic folding.

Another simulation based method was proposed in~\cite{LiIROS2015}.
The authors used an animation software \textit{Autodesk Maya} to evaluate a quality of the fold for a given garment and a folding path.
The folding path was then perturbed until the quality was sufficient.
The model used in~\cite{LiIROS2015} consists of $n$-particles connected by springs.
Spring network structure and the stiffness of individual springs are used to simulates various garment properties.
Such models are computationally effective but tuning the parameters to get real garment behaviour is difficult.
In~\cite{LiIROS2015} author tuned one parameter but the resulting accuracy of the robotic folding was not analysed.

\subsection{Learning Based Approaches}\label{subsec:learningBasedApproaches}
The more recent approaches tried to learn a policy for the folding.
In work~\cite{jia2018learning} the authors proposed a feedback control policy based on the features extracted from an RGBD data.
They proposed a policy which iteratively minimise distance between the current and goal state in a future space.
One of the tested task was a towel folding in which the authors tested a single piece of towel.
Another work~\cite{matas2018sim} learns a convolutional neural network feedback based policy based on RGB image in simulation.
The policy learned in a simulation was then directly evaluated on a single piece of rectangular garment folding in reality.
The learning from demonstration was used in~\cite{sannapaneni2017learning}.
The work was tested on a single rectangular garment.
Deep learning was used in~\cite{yang2017repeatable} and tested on a single piece of rectangular garment too.
A t-shirt was folded in work~\cite{colome2018dimensionality}.
Authors combined learning from demonstration with dimensionality reduction to learn a folding task parametrized by dynamic movement primitives.
A total number of 540~real robot executions were used for training.
All presented learning based approaches solve a task which is more general than a simple folding path design, but
they do not consider the accuracy.
As long as the folding motion imitated the expected folding motion it is considered as success regardless of accuracy of the final garment state.

    \section{FABRIC MODEL}\label{sec:model}
This section describes our model of a fabric strip and introduces the concept of continuation~\cite{wriggers2008nonlinear} which is common in structural mechanics analysis.
We use continuation to analyse the static stability of the strip folding in the subsequent sections.

\subsection{Model Geometry}\label{subsec:modelGeometry}
\begin{figure}[ht]
    \centering
    \subcaptionbox{Reference configuration}[1.0\linewidth]{
    \begin{overpic}[width=1.0\linewidth,tics=10]{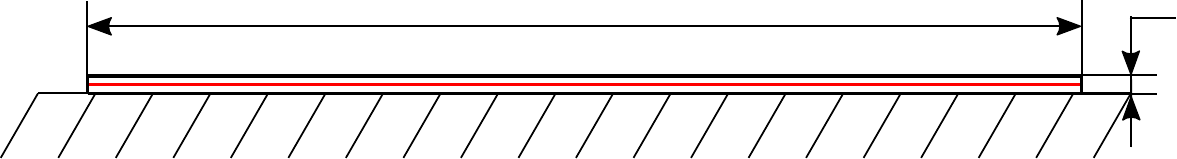}
        \put(50, 13){$l$}
        \put(97, 13){$h$}
    \end{overpic}
    }\\%
    \subcaptionbox{Deformed configuration}[1.0\linewidth]{
    \begin{overpic}[width=1.0\linewidth,tics=10]{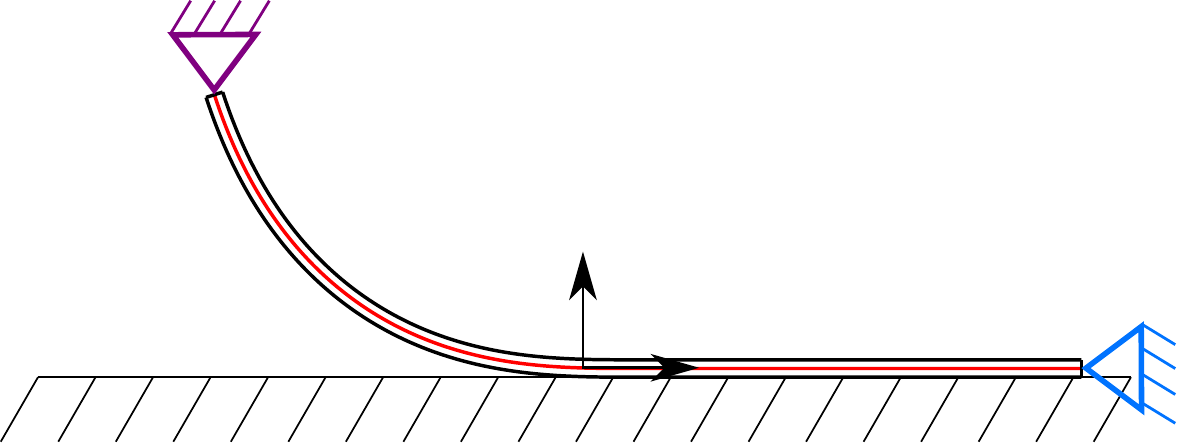}
        \put(58, 8){$x$}
        \put(51, 15){$z$}
    \end{overpic}
    }%
    \caption{
    The state of the model with dimensions~$l \times h \times b$ in the reference~(a) and deformed~(b) configuration.
    The red line represents a mid-surface of the model.
    The position of the model is expressed in a Cartesian coordinate system visualised in the bottom figure.
    The state of the model is constrained by boundary conditions: a frictionless contact between the model and the supporting surface,
    a holding point constraint represented by the blue triangle,
    and a grasping point constraint represented by the purple triangle.
    }
    \label{fig:geometry}
\end{figure}

The strip geometry in reference configuration is represented by a box with dimensions~$l \times h \times b$.
The whole edge with dimension~$b$ is grasped and manipulated during the folding and the fabric material is assumed to be homogeneous and isotropic.
Consequently, the model can be represented by 2D~geometry as shown in Fig.~\ref{fig:geometry}.
We refer our model to a Cartesian system placed in the middle of the strip~(Fig.~\ref{fig:geometry}).
When visualising the model state, only the mid-surface (i.e.\ a cut $z=0$ in a reference configuration) is plotted.

\subsection{Boundary Conditions}\label{subsec:boundaryConditions}
During the folding, three boundary conditions are considered~\cite{PetrikADR2017}:
(i) frictionless contact between the supporting surface and the fabric,
(ii) a holding point condition which prevents the strip from slipping, and
(iii) a grasping point condition which simulates the gripper grasping.
All these boundary conditions are visualised in Fig.~\ref{fig:geometry}.

\subsection{Constitutive Relation}\label{subsec:constitutiveRelation}
The fabric strip model is represented by a 3D elastic solid.
A stress-strain relation needs to be specified to describe an elastic solid.
This constitutive relation characterizes the behaviour of a material.

As a strain measure, we use the Green-Lagrange strain~\cite{wriggers2008nonlinear} defined as:
\begin{align}
    \bm E = \frac{1}{2} \left( \bm F^\top \bm F - \bm I \right), \quad \bm F = \bm I + \nabla \bm u,
\end{align}
where $\bm F$ denotes a deformation gradient, $\bm I$ stands for an identity tensor and $\bm u$ represents the displacement of the solid from its reference configuration.
The Second Piola-Kirchhoff stress tensor~$\bm S$~\cite{wriggers2008nonlinear} is usually used in the presence of large deformations.
It is computed as:
\begin{align}
    \bm S = \det \bm F \, \bm F^{-1} \left( \bm C : \bm E \right) \bm F^{-\top},
\end{align}
where $\bm C$ is the elastic tensor and the symbol~`$:$' stands for the double-dot tensor product.

An isotropic homogeneous material is assumed.
In this case, the elastic tensor depends on Young's modulus~$E$ and Poisson's ratio~$\nu$ only.
Because of the tensor symmetry we can represent the double-dot product $\bm C : \bm E$ completely in a Voigt notation~\cite{wriggers2008nonlinear}:
\begin{align}
    \bm \sigma &= \frac{E}{\left( 1+\nu \right)\left( 1-2\nu \right)} \, \bm D \, \bm \varepsilon \, ,\\
    \bm D &= \left( \begin{matrix}
                        1-\nu & \nu & \nu & 0 & 0 & 0 \\
                        \nu & 1-\nu & \nu & 0 & 0 & 0 \\
                        \nu & \nu & 1-\nu & 0 & 0 & 0 \\
                        0 & 0 & 0 & \frac{1-2 \nu}{2} & 0 & 0 \\
                        0 & 0 & 0 & 0 & \frac{1-2 \nu}{2} & 0 \\
                        0 & 0 & 0 & 0 & 0 & \frac{1-2 \nu}{2} \\
    \end{matrix} \right) ,
\end{align}
where $\bm \varepsilon$ is the Voigt form of the strain tensor~$\bm E$ and $\bm \sigma$ is the result of the double-dot product in the Voigt form.

\subsection{Relation to \textit{weight-to-stiffness-ratios}}\label{subsec:relationToEtas}
A Poisson's ratio~$\nu$, \textit{weight-to-membrane-stiffness-ratio}~$\eta_m$, and \textit{weight-to-bending-stiffness-ratio}~$\eta_b$ were used in~\cite{PetrikICRA2018} to describe the strip model.
These parameters model the elastic membrane stiffness independently on the bending stiffness.
The \textit{weight-to-stiffness-ratios} are computed as:
\begin{align}
    \eta_m = {\left( 1-\nu^2 \right)}\frac{\rho}{h \, E}, \qquad \eta_b = {12\left( 1-\nu^2 \right) }\frac{\rho}{h^3 \, E}\, \text{.}
\end{align}
We use these equations to map the~\textit{weight-to-stiffness-ratios} into the model described in Sec.~\ref{subsec:constitutiveRelation}.
Note that by using this mapping the parameter~$h$ no longer represents the physical thickness of the strip.
Furthermore,~$\nu$ and~$\eta_m$ do not have significant impact on the folding path because the fabric is usually not stretchable significantly~\cite{PetrikICRA2018}.
In our experiments, we fixed these parameters to values:
\begin{align}
    \nu = 0 \, , \qquad \eta_m = 10^{-3} \text{~\etamunit} \, \text{.}
\end{align}

\subsection{Equilibrium Equation}\label{subsec:equilibrium}
The equilibrium equation is given by Newton's second law:
\begin{align}
    \label{eq:dynEquilibrium}
    \rho \frac{\partial^2 \bm u}{\partial t^2} = \nabla \cdot \left( \bm F \bm S \right)^\top + \rho \, \bm g \, ,
\end{align}
where $\rho$ is material density, $t$ represents time, and $\bm g$ is the gravitational acceleration vector.
In case of a static equilibrium problem, the inertial term given by the left side of Eq.~\eqref{eq:dynEquilibrium} can be neglected:
\begin{align}
    \label{eq:statEquilibrium}
    \bm 0 = \nabla \cdot \left( \bm F \bm S \right)^\top + \rho \, \bm g \, \mathrm{.}
\end{align}

Eq.~\eqref{eq:dynEquilibrium} or~\eqref{eq:statEquilibrium} together with boundary conditions forms a strong form~\cite{wriggers2008nonlinear} of the problem.
In finite element modeling, the strong form is transformed into the weak form~\cite{wriggers2008nonlinear} which is then expressed by the nonlinear residual equation
\begin{align}
    \label{eq:residualForce}
    \bm r ( \bm u , \lambda ) = 0,
\end{align}
where $\bm r $ is a residual force vector and $\lambda$ is a single control parameter, e.g.~a pseudo time or a gripper force holding the strip.
In a static case, the residual force represents the balance of internal and external forces while in a dynamic case there is an additional contribution from inertial terms.
For a fixed control parameter~$\lambda$, Eq.~\eqref{eq:residualForce} can be solved iteratively using Newton-Raphson method~\cite{wriggers2008nonlinear}
\begin{align}
    \label{eq:newtonRaphson}
    \bm K ( \bm u , \lambda ) \Delta \bm u = - \bm r ( \bm u , \lambda ),
\end{align}
where $\bm K ( \bm u , \lambda ) $ is called stiffness matrix for a static problem.
Hereinafter, the static problem is assumed unless explicitly stated otherwise.

\subsection{Continuation}\label{subsec:continuation}
Changing parameter $\lambda$ incrementally and seeking for a state in equilibrium for each value of $\lambda$ is called continuation.
The state in equilibrium satisfies Eq.~\eqref{eq:residualForce} and the sequence of equilibrium states for a varying value of control parameter is called equilibrium path.

Two phenomena are observed during continuation in structural mechanics: snap-through (limit point) and buckling (bifurcation).
In a snap-through case, the increasing value of control parameter at some point~$\lambda_{c}$ causes the system to change its configuration significantly, i.e.~there is a `jump' in a state vector $\bm u$.
The buckling means that there is a point~$\lambda_c$ at which two equilibrium paths intersect.
The real system chooses which path is followed.
The simulated ideal system is not able to decide the path solely from Eq.~\eqref{eq:newtonRaphson}.
Point~$\lambda_c$ is called a critical point in both cases.

At the critical point the stiffness matrix becomes singular.
The common indicator used to detect the singularity is the value of the smallest eigenvalue of the stiffness matrix.
At the critical point the smallest eigenvalue is zero.
The structural mechanics interpretation of the critical point is that there exists a non zero displacement~$\Delta \bm u$ which requires no additional force.
The direction of this displacement is given by the eigenvector associated with the zero eigenvalue.
Consequently, the system is not statically stable.

    \section{CONTINUATION IN X-AXIS}\label{sec:left2right}
\begin{figure}[ht]
    \centering
    \subcaptionbox{Initial state~$\lambda_0$\label{fig:l2ra}}[0.5\linewidth]{\includegraphics[width=0.5\linewidth]{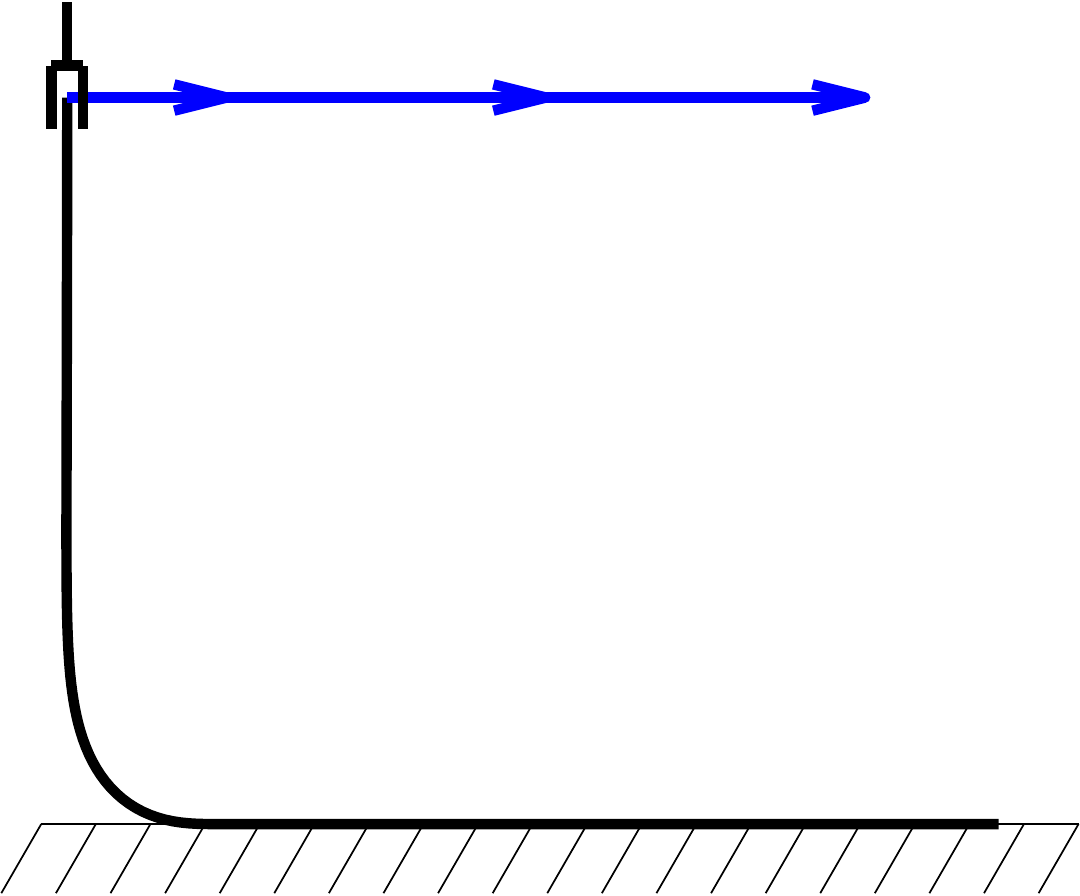}}%
    \subcaptionbox{State near a critical point~$\lambda_{c}$\label{fig:l2rb}}[0.5\linewidth]{\includegraphics[width=0.5\linewidth]{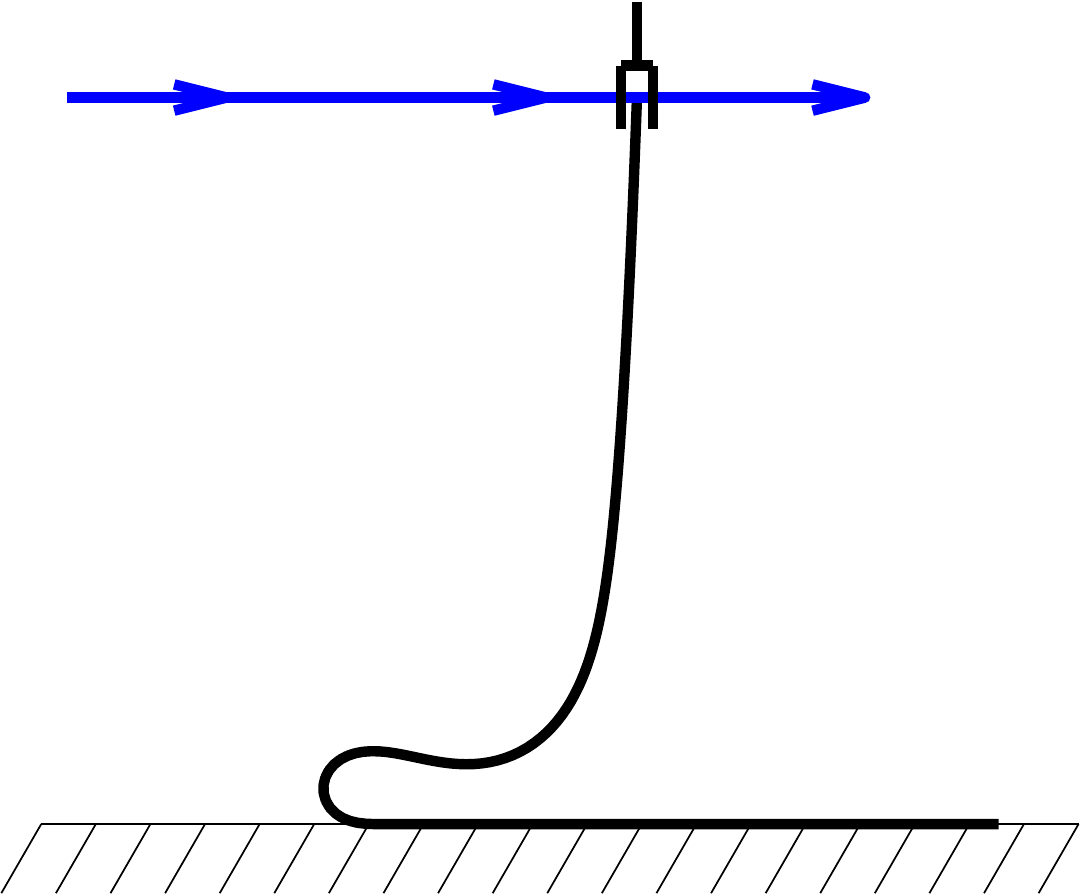}}%
    \caption{
        The continuation of the control parameter~$\lambda$ which corresponds to the gripper position in x-axis.
        The continuation starts from the value~$\lambda_0$ and the direction of continuation is shown by the blue arrows~(a).
        The value of the control parameter is increasing until the solver failed to converge, which is a consequence of a singular stiffness matrix.
        The value of~$\lambda$ at which the stiffness matrix is singular is called a critical point~$\lambda_c$~(b).
    }
    \label{fig:l2r}
\end{figure}

\begin{figure}[ht]
    \centering
    \includegraphics[width=0.8\linewidth]{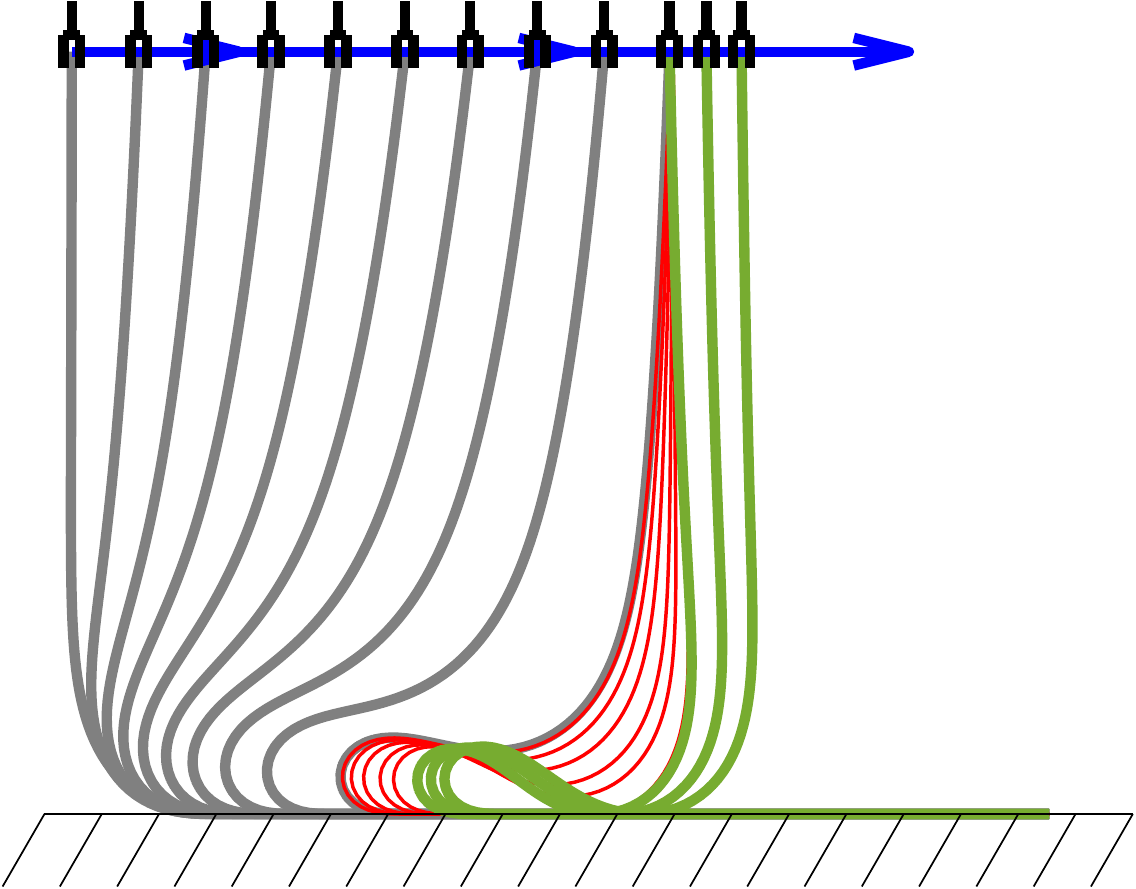}
    \caption{
        The strip states found by continuation starting from~$\lambda_0$ are shown in grey.
        After the critical point~$\lambda_{c}$ is reached, the dynamical solver is used to trace the state evolution and the result is shown in red.
        The dynamical solver is stopped after the strip touches the ground.
        Then the strip is statically stable and the continuation continues as shown by states in green.
    }
    \label{fig:l2rdyn}
\end{figure}

\begin{figure}[ht]
    \centering
    \includegraphics[width=.8\linewidth]{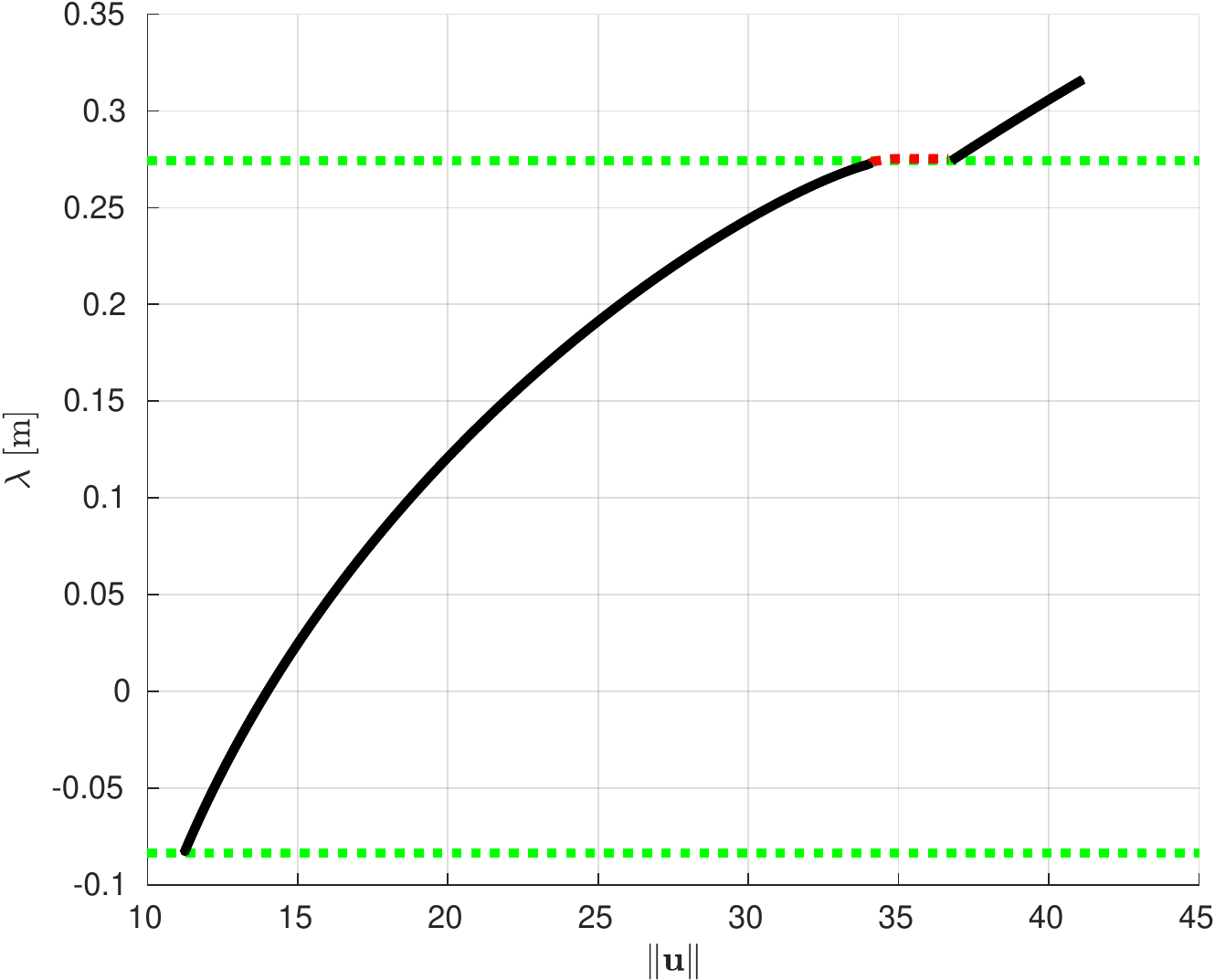}~%
    \caption{
        The control parameter/deflection curve.
        The black part of the curve was obtained by continuation in static solver.
        The red part was found by the dynamical solver.
        The green lines show values of $\lambda_0$(bottom) and $\lambda_c$(top).
    }
    \label{fig:l2rlambdaDeflection}
\end{figure}

This section studies the fabric strip motion induced by varying gripper position in x-axis.
Therefore, in this case the parameter~$\lambda$ represents the gripper position in x-axis.
The static stability is analysed according to Section~\ref{subsec:continuation}.
The start state of the strip and the studied motion are visualised in Fig.~\ref{fig:l2ra}.
The value of the control parameter is increased starting from value~$\lambda_0$.
At the critical point~$\lambda_c$, the solution of Eq.~\eqref{eq:newtonRaphson} can not be found due to the singular stiffness matrix.
From this point onwards it is not possible to find consequent strip states by static analysis because the strip starts to move dynamically.
However, the strip states can be simulated using a dynamic solver.
After the strip touches the ground, the problem becomes statically stable and we can switch back to the static continuation.
The states found by this approach are visualised in Fig.~\ref{fig:l2rdyn}.
Another visualisation of the phenomena plots the control parameter/deflection curve as shown in Fig.~\ref{fig:l2rlambdaDeflection}.

\begin{figure}[ht]
    \centering
    \subcaptionbox{Paths performed in simulation}[.8\linewidth]{\includegraphics[width=.8\linewidth]{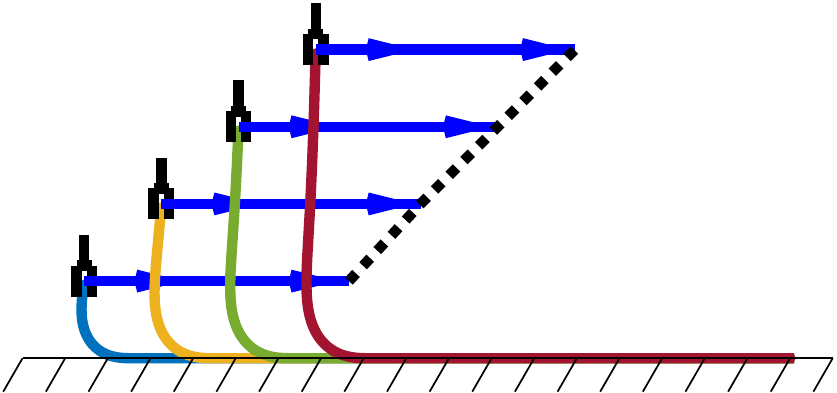}}\\%
    \subcaptionbox{Paths performed in reality}[.8\linewidth]{\includegraphics[width=.8\linewidth]{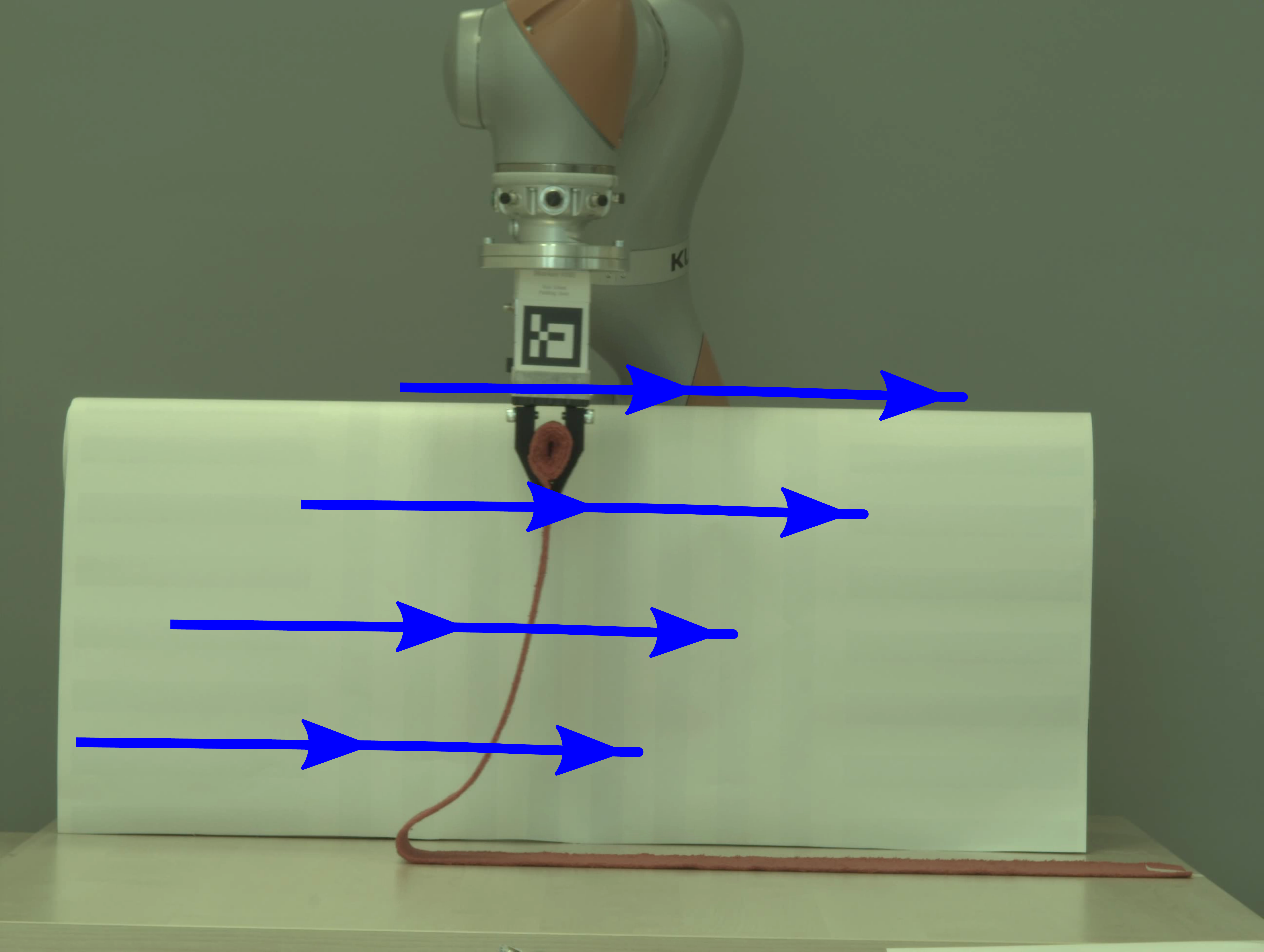}}~%
    \caption{
        The paths for estimating values of critical points for different gripper height.
        In model~(a) the continuation for the given height stops at the critical point.
        The black dotted line shows the values of~$\lambda_c$.
        In reality~(b) the gripper motion continues and the values of critical points are estimated manually offline.
        The robot and camera were geometrically calibrated.
    }
    \label{fig:l2rCriticalPointsMeasurement}
\end{figure}

\begin{figure}[!ht]
    \centering
    \includegraphics[width=.8\linewidth]{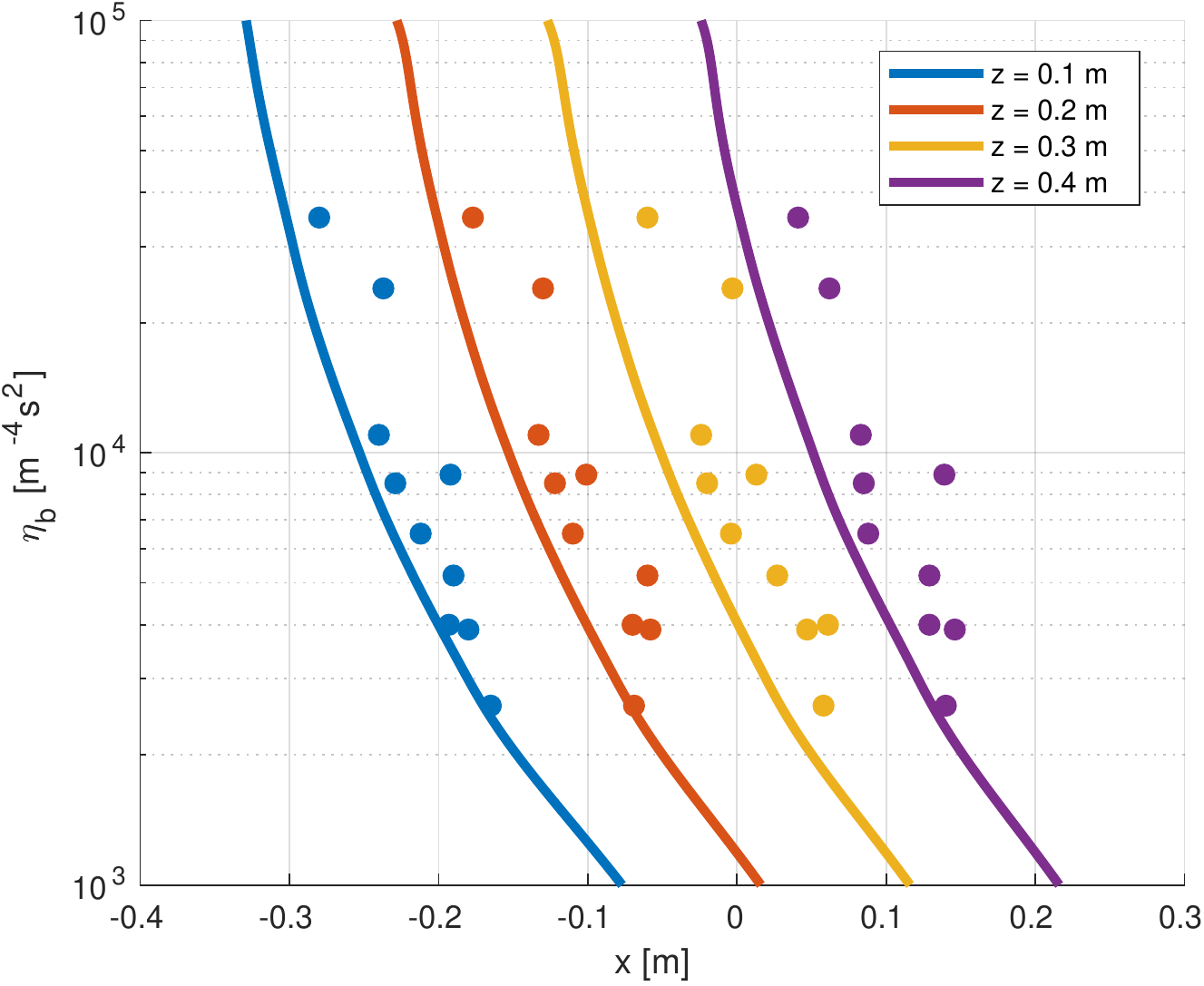}~%
    \caption{
    The observed values of~$\lambda_c$ for different gripper heights~$z$ for modelled~(solid line) and for real~(dots) strips.
    The parameter~$\eta_b$ represents the \textit{weight-to-bending-stiffness-ratio} and for the real strips it was estimated from the maximal height of the folded state.
    }
    \label{fig:l2rRealLambda}
\end{figure}

We observed the static stability loss phenomena while manipulating real fabric strip as well.
To compare modelled and real observations we evaluate a value of $\lambda_c$ for several gripper heights as shown in Fig.~\ref{fig:l2rCriticalPointsMeasurement}.
The modelled and real values of $\lambda_c$ are shown in Fig.~\ref{fig:l2rRealLambda}.

The real measurement was performed for several fabric strips with varying materials.
It can be seen in Fig.~\ref{fig:l2rRealLambda} that the modelled values of critical points underestimate the real values.
Moreover, the difference between these values varies for different types of the fabric strip.
This observation can be attributed to properties which were omitted in the modelling process such as internal friction between individual fabric threads and nonlinear stiffness.
The simplification makes computation of the model state tractable but the modelled and real states differ slightly as the model reaches the singularity.

On the other hand, the fact that the model always underestimates the value of critical point can be used for planning.
If the strip following a given path is stable in the model then it will be stable also in reality regardless of the value of omitted parameters.
This observation will be used in the assessment of state-of-the-art folding paths in the next section.

    \section{ASSESSMENT OF FOLDING PATHS}\label{sec:foldingPathsAssessment}
We next study folding paths proposed in literature using the static analysis.
In this section the control parameter~$\lambda$ represents a pseudo time.
The gripper position is given by a folding path and depends on the control parameter.

\begin{figure*}[t]
    \centering
    \subcaptionbox{\textit{R-path}\label{fig:pathstatesR}}[.33\linewidth]{\includegraphics[width=.33\linewidth]{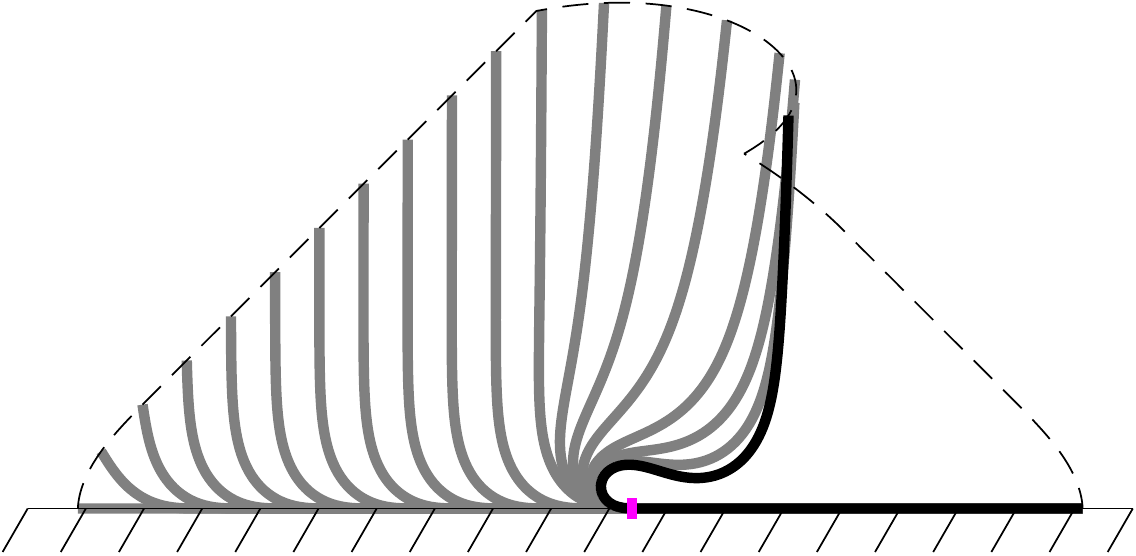}}%
    \subcaptionbox{\textit{Triangular path}\label{fig:pathstatesT}}[.33\linewidth]{\includegraphics[width=.33\linewidth]{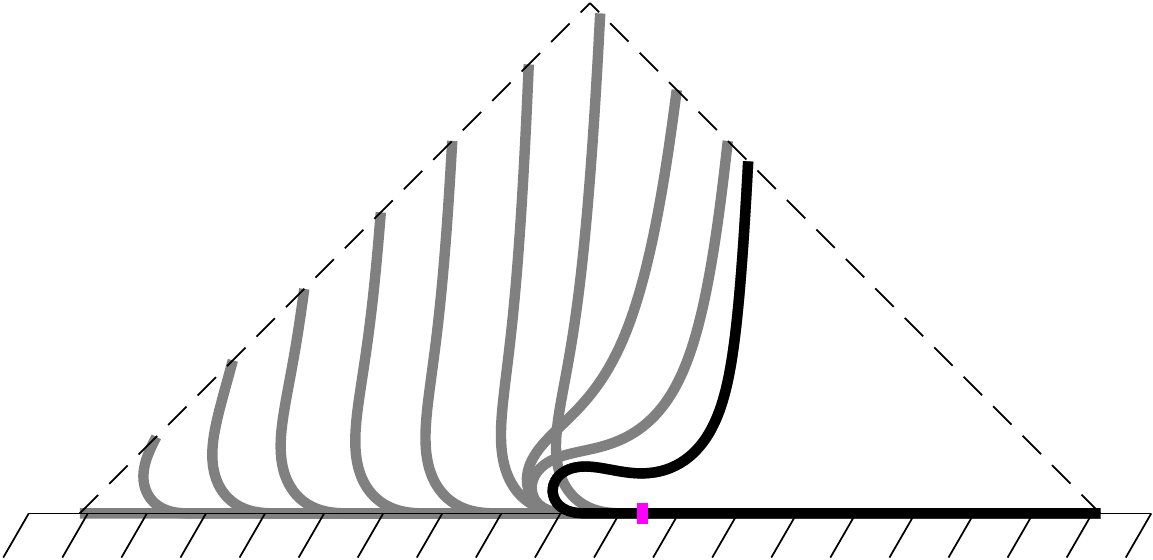}}%
    \subcaptionbox{\textit{Circular path}\label{fig:pathstatesC}}[.33\linewidth]{\includegraphics[width=.33\linewidth]{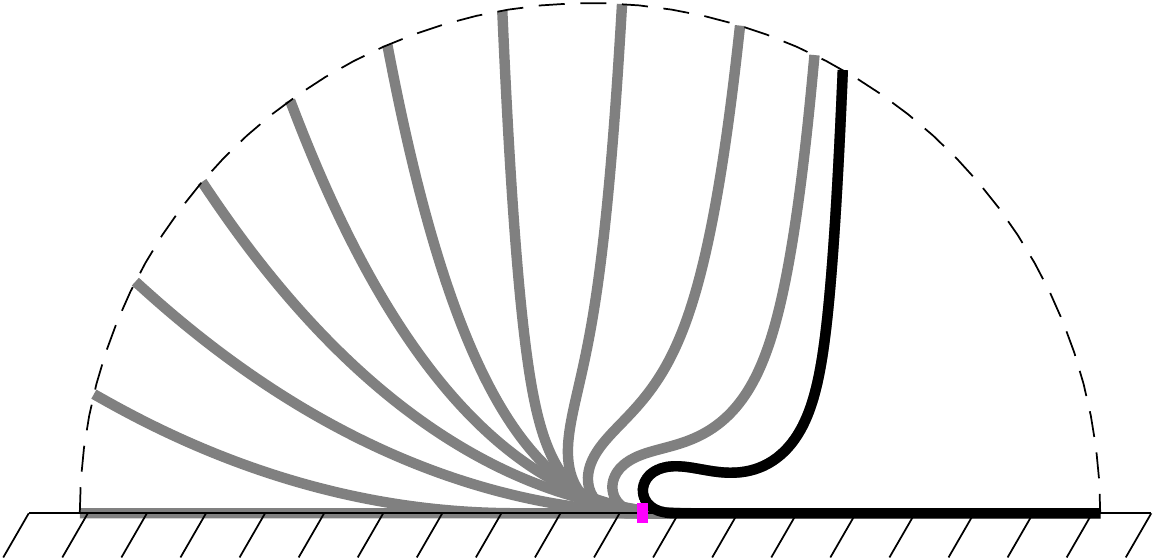}}%
    \caption{
    The fabric state evolution when following different folding paths are shown in gray.
    The state in black shows the last state found before a critical point is reached.
    A magenta color indicates the position at which the strip touches the ground in the folded state,
    i.e.\ in the folded state, the part of the strip between this point and the holding point lies on the ground.
    }
    \label{fig:pathstates}
\end{figure*}

\subsection{R-path}\label{subsec:r-path}
\begin{figure}[ht]
    \centering
    \subcaptionbox{Dynamic motion\label{fig:criticalstatesRa}}[.5\linewidth]{\includegraphics[width=.5\linewidth]{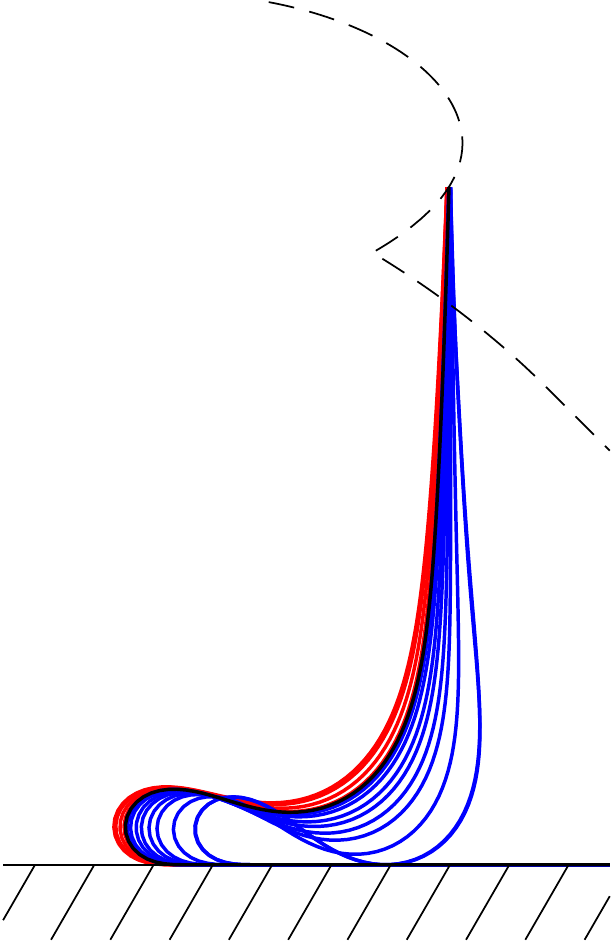}}%
    \subcaptionbox{Static states\label{fig:criticalstatesRb}}[.5\linewidth]{\includegraphics[width=.5\linewidth]{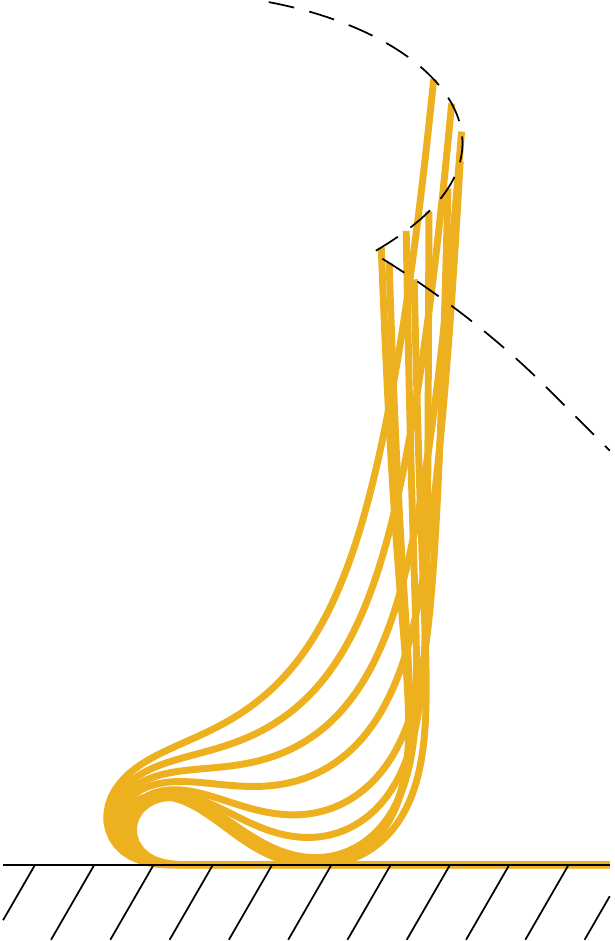}}%
    \caption{
    The states evolution when passing critical point without internal friction~(a) and with internal friction~(b).
    The dynamic solver was used to trace red and blue states~(a).
    }
    \label{fig:criticalstatesR}
\end{figure}

Consider a continuation of the control parameter for a folding path called~\textit{R-path}~\cite{PetrikADR2017}.
The continuation is stopped when a critical point is reached as shown in Fig.~\ref{fig:pathstatesR}.
At this point, the dynamical solver can be used to trace the state evolution.
We observed that there are two different directions which the strip can dynamically follow as shown in Fig.~\ref{fig:criticalstatesRa} by red and blue color.
In simulation we used a disturbance to select one of the directions.

\begin{figure}[ht]
    \centering
    \includegraphics[width=.8\linewidth]{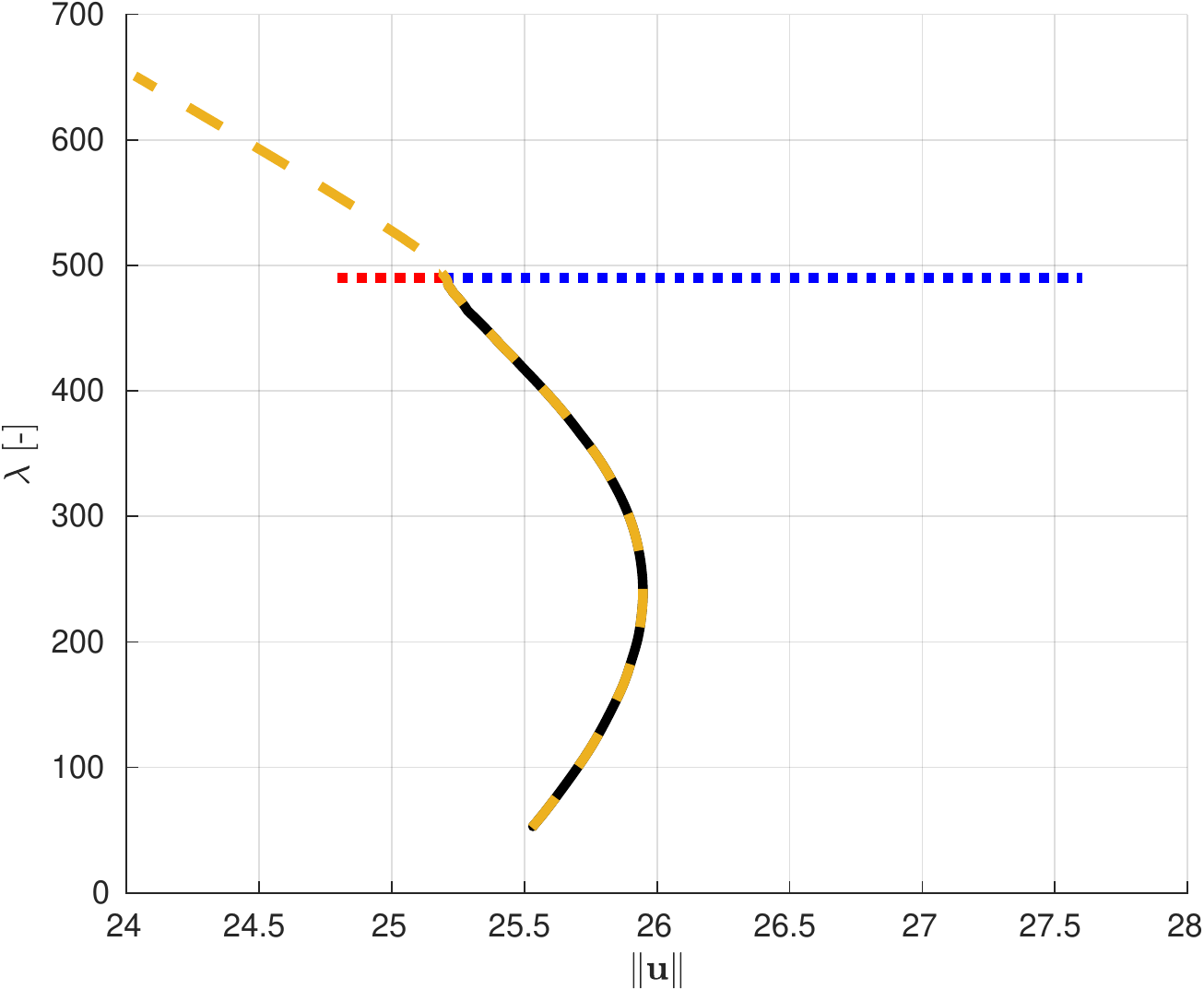}~%
    \caption{
    Control parameter/displacement curve when following \textit{R-path}.
    Black color represents the continuation before the critical point is reached.
    At the critical point, the dynamic solver was used to track the red or blue curve.
    The yellow color represents continuation in a case with simulated internal friction.
    }
    \label{fig:pathdisplacementcurve}
\end{figure}

However, in Section~\ref{sec:left2right} we observed that in reality the strip remains stable longer than predicted by the model.
This can be attributed to internal friction.
We approximate it by a force which pulls the current state of the model towards the previous state.
The pulling force was a few orders of magnitude smaller than the gripper force.
With this model we are able to follow the whole \textit{R-path} as shown in Fig.~\ref{fig:criticalstatesRb}.
The control parameter/deflection curve is shown in Fig.~\ref{fig:pathdisplacementcurve}.

The added internal friction stabilizes the model and the folding path could be followed.
With the internal friction, the uncontrolled dynamical motion is avoided and the states of the strip correspond to the planned states.
Consequently, the position of point where the strip touches itself for the first time equals to its expected position.
This was experimentally confirmed in~\cite{PetrikIROS2016,PetrikMESAS2016,PetrikADR2017}.

\subsection{Triangular and Circular Path}\label{subsec:triangularAndCircularPath}
\textit{Triangular path} was designed for an infinitely flexible material (i.e.~$\eta_b = \infty$).
If the path is used for the material with finite flexibility, it reaches the critical point as shown in~Fig.~\ref{fig:pathstatesT}.
Then it falls down dynamically similarly to Fig.~\ref{fig:l2rdyn}.
However, in an ideal model the position at which the strip touches itself does not correspond to the planned position because the flexibility is different.
The $x$-coordinate of the first touch position is smaller than planned.
For real fabric material, the position at which the strip falls down is hard to predict according to our measurements shown in Section~\ref{sec:left2right}.
For one particular material, the position at which the strip touches itself would be correct.

For a \textit{circular path}, the situation is similar to the \textit{triangular path} as shown in Fig.~\ref{fig:pathstatesC}.
The strip will fall down dynamically.
However, the $x$-coordinate of the first touch position is larger than planned.
For real fabric material the $x$-coordinate of this position is even larger according to our observations from Section~\ref{sec:left2right}.

\subsection{Discussion}
For \textit{triangular} and \textit{circular paths}, snap-through behaviour occurs after the critical point is reached.
The critical point is underestimated in our model, i.e.\ the snap-through behaviour occurs later than predicted.
Thus, outcome of both paths is hardly predictable by a model and requires dynamical modeling.

On the other hand, the \textit{R-path} avoids the snap-thorugh behaviour if there is internal friction in the system.
This friction corresponds to the observation that the model underestimates the value of critical point for fabric material.
The following of the \textit{R-path} is statically stable for materials with internal friction.

The \textit{R-path} has an advantage compared to the other folding paths:
At the beginning, it rise up the strip until the part of the strip which lies on the ground is same as in the folded state as shown in Fig.~\ref{fig:pathstates}.
The rest of the path does not change the amount of the strip which lies on the ground.
The path does not rely on the snap-through outcome and its only assumption is that the strip remains stable.

The existence of the force which prevents the uncontrolled motion of the strip is not limited to fabric material only.
The \textit{R-path} was successfully used to accurately fold rubber strips in~\cite{PetrikIROS2016}.
The origin of the force in rubber material is different than thread friction and may correspond e.g.\ to memory effect of the material.
However, there are materials without this force, for example thin metal.
Using \textit{R-path} for folding this material would not be feasible because any disturbance in a gripper position would cause the dynamic motion.
For this material it may be better to plan a path which utilizes the snap-through behaviour.
Predicting the outcome of the dynamic motion would be possible for material without the internal friction.

    \section{CONCLUSION}\label{sec:conclusion}
The paper presented a static instability analysis for a single layer strip folding.
It was shown that the instability exists in the folding process and that it results in the dynamic motion of the strip.
The presented comparison of the real fabric strip and the simulated model suggested that there is an internal friction in the real system.
This observation was used to analyse the state-of-the-art folding paths in a presence of the static instability.
It was shown that \textit{R-path} folds a strip accurately for materials with internal friction and that another path may be more accurate for other materials.

This paper studied only strip folding.
However, the same method can be used to predict static instability for folding garments of more complicated shapes.
For example, it can be used to predict a collapse of a single arm rectangular garment folding which was shown in~\cite{PetrikADR2017}.
The detected instability can be then considered in planning to avoid the region of instability or to immobilize the garment with another gripper.
The eigenanalysis of the stiffness matrix can likely be used to determine the position of the additional gripper.

The concept of static instability might also be used in areas not related to folding.
For example in grasping of a deformable object the instability would suggest that the object is not held properly.
Again the eigenanalysis can reveal the part of object which is not stable.

    \bibliographystyle{IEEEtran}
    \bibliography{bib/IEEEabrv,bib/mine,bib/tmech_stability}

\end{document}